\documentclass[letterpaper, 10pt, conference]{ieeeconf}
\IEEEoverridecommandlockouts 
\usepackage[dvipsnames,table]{xcolor}

\usepackage{graphicx}
\usepackage{setspace}
\usepackage{amsmath, amsfonts, amssymb}
\usepackage{cleveref}
\usepackage{booktabs}
\usepackage{multirow}
\usepackage{float}

\title{COARSE: Collaborative Pseudo-Labeling with Coarse Real Labels for Off-Road Semantic Segmentation}

\author{
Aurelio Noca$^{1}{^,}^{2}$, Xianmei Lei$^{1}$, Jonathan Becktor$^{1}$, Jeffrey Edlund$^{1}$,  Anna Sabel$^{1}$, \\ Patrick Spieler$^{1}$, Curtis Padgett$^{1}$, Alexandre Alahi$^{2}$, Deegan Atha$^{1}$ 
\thanks{*The research was carried out at the Jet Propulsion Laboratory, California Institute of Technology, under a contract with the National Aeronautics and Space Administration (80NM0018D0004). This work was partially supported by the Defense Advanced Research Projects Agency (DARPA). The High Performance Computing resources used in this investigation were provided by funding from the JPL Information and Technology Solutions Directorate. © 2025. All rights reserved.}
\thanks{$^{1}$Jet Propulsion Laboratory, California Institute of Technology,
4800 Oak Grove Dr, Pasadena, CA 91011.}%
\thanks{$^{2}$EPFL, Ecole Polytechnique Federale de Lausanne, Rte Cantonale, 1015 Lausanne, Switzerland.}%
}

\begin{document}

\maketitle



\begin{abstract}


Autonomous off-road navigation faces challenges due to diverse, unstructured environments, requiring robust perception with both geometric and semantic understanding. However, scarce densely labeled semantic data limits generalization across domains. Simulated data helps, but introduces domain adaptation issues. We propose COARSE, a semi-supervised domain adaptation framework for off-road semantic segmentation, leveraging sparse, coarse in-domain labels and densely labeled out-of-domain data. Using pretrained vision transformers, we bridge domain gaps with complementary pixel-level and patch-level decoders, enhanced by a collaborative pseudo-labeling strategy on unlabeled data. Evaluations on RUGD and Rellis-3D datasets show significant improvements of 9.7\% and  8.4\% respectively, versus only using coarse data. Tests on real-world off-road vehicle data in a multi-biome setting further demonstrate COARSE’s applicability.

\end{abstract}

\section{Introduction}

Off-road navigation presents itself as a challenge in many applications, including agricultural automation, disaster response and planetary exploration. Traversing unstructured and highly varied terrain -- featuring unpredictable slopes, dense vegetation, and hidden obstacles -- poses unique challenges that even skilled human drivers struggle to navigate reliably.
Although standard approaches for mapping and geometric traversability analysis  provide a framework for obstacle avoidance, they fail to distinguish between obstacles that are genuinely solid and those that appear solid but are actually traversable (e.g., tall grass, bushes, water puddles). Consequently, accurate semantic segmentation is a key requirement for safe and robust off-road autonomy.

\begin{figure}[t] 
\centering
\includegraphics[width=\linewidth]{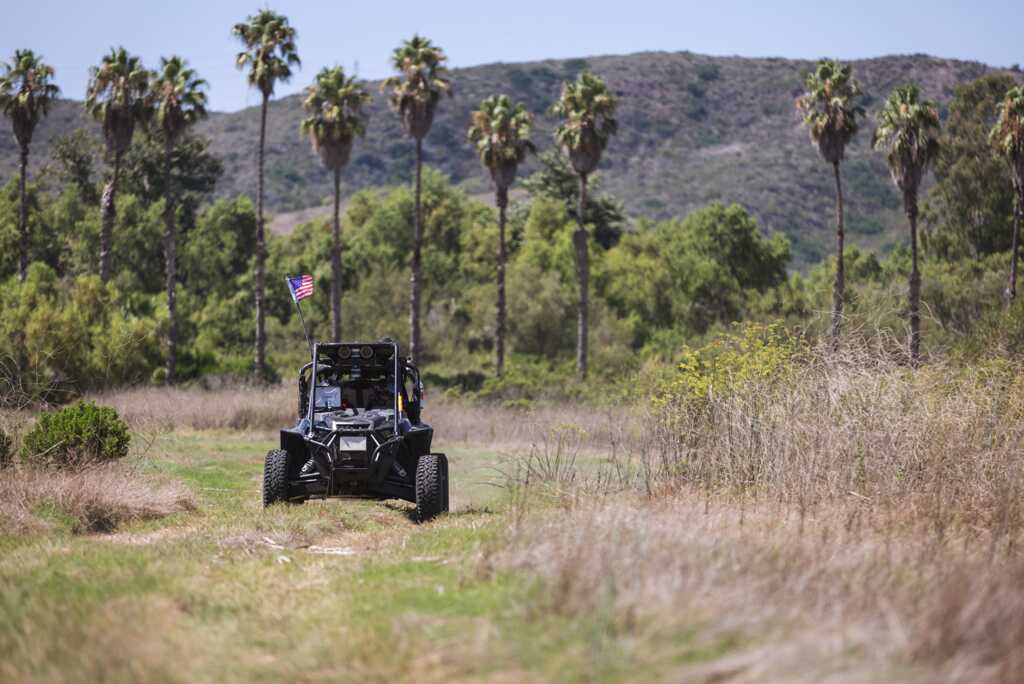}
\caption{Polaris RZR dune buggy, outfitted with an extensive sensor suite for autonomous off-road navigation, being driven in the San Diego Grasslands biome.}
\label{fig:vehicle}
\end{figure}

At the same time, the scarcity of densely labeled data in off-road domains limits the applicability of deep learning-based segmentation models. Producing large-scale pixel-wise annotations is laborious and expensive, especially when many environments exhibit ambiguous boundaries and overlapping vegetation classes. Recent work has thus moved toward semi-supervised learning (SSL) and domain adaptation techniques \cite{zou2019confidence, berthelot2019mixmatch}, leveraging unlabeled data alongside sparse ground-truth labels. Moreover, recent vision foundation models \cite{Caron2021EmergingPI, Oquab2023DINOv2LR} have demonstrated strong cross-domain generalization, mitigating some of the burden of in-domain labeling. In this paper, we introduce COARSE, a novel semi-supervised domain adaptation framework that fuses \emph{coarse in-domain labels} with \emph{densely annotated out-of-domain} data in a collaborative pseudo-labeling pipeline, featuring complementary pixel-level and patch-level decoders. We validate our approach on publicly available off-road datasets and demonstrate its effectiveness for real-world autonomous driving scenarios (see \cref{fig:vehicle}).

\begin{figure*}[h!]
    \centering
    \includegraphics[width=0.7\linewidth]{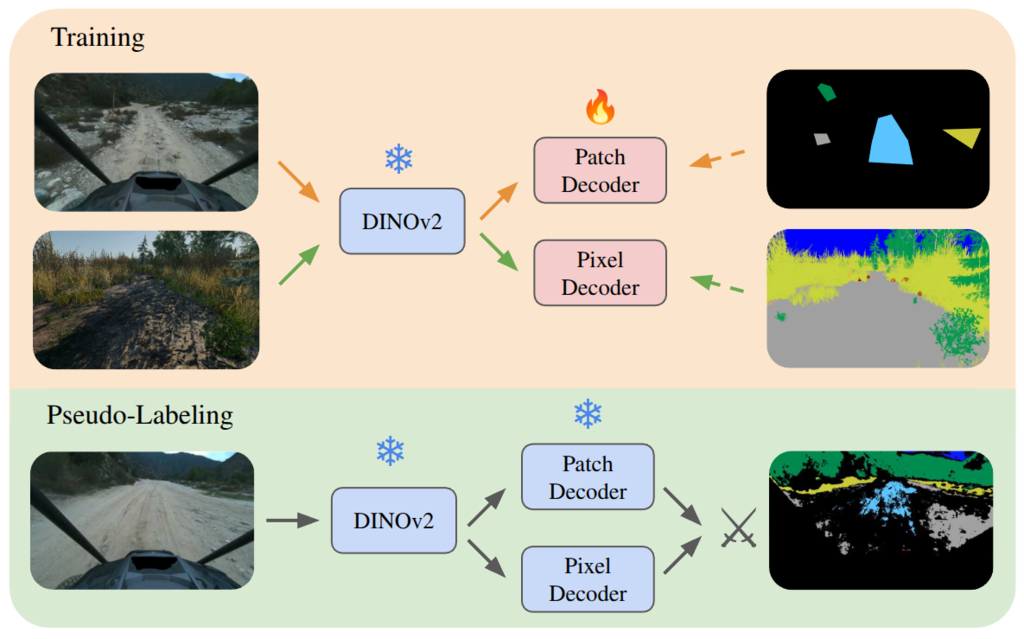}
    \caption{Our COARSE pseudo-labeling approach leverages two decoders -- PatchDecoder and PixelDecoder -- both utilizing the robust semantic features of the DINOv2 encoder. The PixelDecoder further integrates low-level geometric details from the input image. The PatchDecoder is trained on coarse ID data, while the PixelDecoder is trained on a combination of dense OOD and coarse ID data. Pseudo-labels are generated via disagreement of predicted semantic maps.}
    \label{fig:method}
\end{figure*}

\section{Related Work}
Early off-road segmentation methods often relied on handcrafted features such as color-space thresholding or texture filters to delineate vegetation from drivable surfaces~\cite{stanley_darpa_grand_challenge}. Although these approaches worked well in controlled conditions, they generalized poorly to different terrains. With the success of convolutional neural networks (CNNs), fully convolutional architectures like FCN \cite{Shelhamer2014FullyCN} and U-Net \cite{Ronneberger2015UNetCN} paved the way for end-to-end semantic segmentation. More recent methods have incorporated attention-based models (e.g., Segmenter \cite{Strudel2021SegmenterTF}, SegFormer \cite{Xie2021SegFormerSA}, DPT \cite{Ranftl2021VisionTF}, and Mask2Former \cite{Cheng2021MaskedattentionMT}), showing improved accuracy on dense labeling tasks. Additionally, investigations into ViTs models have demonstrated their resilience to distribution shifts and image perturbations like illuminations changes and noise \cite{Bhojanapalli2021UnderstandingRO,Naseer2021IntriguingPO,Paul2021VisionTA,Zhou2022UnderstandingTR}.

However, obtaining extensive dense labels for off-road environments remains prohibitively expensive. Benchmarks such as RUGD \cite{Wigness2019ARD} and Rellis-3D \cite{Jiang2020RELLIS3DDD} illustrate large class imbalances and unclear boundaries, creating a strong need for data-efficient approaches. Semi-supervised learning (SSL) leverages large amounts of unlabeled data to compensate for limited annotations \cite{zou2019confidence, berthelot2019mixmatch, Becktor2022-5}, employing strategies like consistency regularization \cite{tarvainen2017mean} or pseudo-labeling \cite{lee2013pseudo}. Meanwhile, domain adaptation methods tackle the shift between labeled source data (e.g., synthetic or out-of-distribution images) and unlabeled target data \cite{tzeng2017adversarial, long2015learning}.

A key challenge in off-road segmentation is label scarcity and noise due to ambiguity, as many scenes can be coarsely or partially annotated. Recent work has shown that pretrained ViTs enable using coarse labels for few-shot semantic segmentation across multiple domains \cite{Atha2024FewshotSL}. Alternatively, coarse labels can be refined via test-time augmentation \cite{Das2022UrbanSS}, or limited real data can be enhanced by synthetic data \cite{Tranheden2020DACSDA, Wang2021DomainAS,Becktor2022-2,Becktor2022-3, grauer2024visionbaseddetectionuncooperativetargets}. Further work has explored the use of a dual model architecture, a CNN and a Transformer, where each model's output pseudo-labels are used to train the other \cite{Luo2021SemiSupervisedMI}.


Large pretrained vision encoders have emerged as a crucial building block in computer vision. Starting from earlier backbones like ResNet \cite{He2015DeepRL} or EfficientNet \cite{Tan2019EfficientNetRM}, modern self-supervised models such as MAE \cite{He2021MaskedAA} and DINO \cite{Caron2021EmergingPI} demonstrate robustness and adaptability with minimal labeled data, culminating in DINOv2 \cite{Oquab2023DINOv2LR}, which excels across a wide range of tasks. DINOv2 has further shown great promise in producing domain-independent features. By exploiting these pretrained features, an off-road segmentation pipeline can converge faster and achieve superior generalization. COARSE builds on these insights, using DINOv2 as a feature extractor followed by a complementary pair of decoders and large-scale unlabeled data to tackle off-road semantic segmentation with sparse real labels.



\section{Method}

\subsection{Datasets}
\label{sec:method_datasets}

\begin{figure}[h!]
    \centering
    \includegraphics[width=\linewidth]{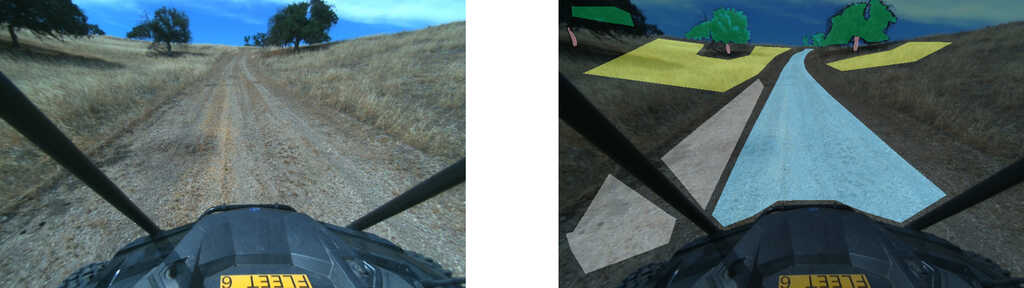}
    \includegraphics[width=\linewidth]{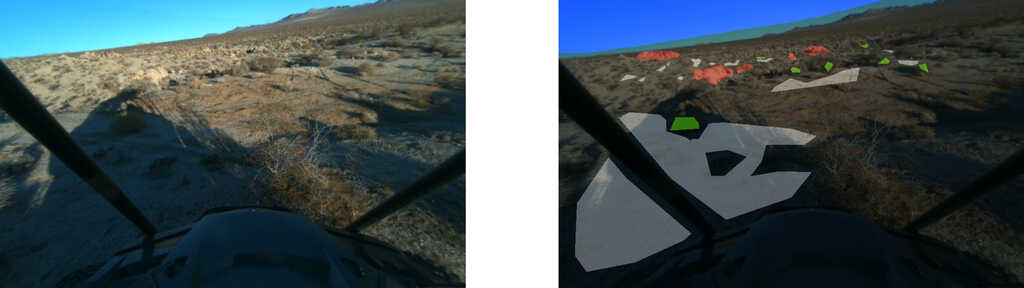}
    \includegraphics[width=\linewidth]{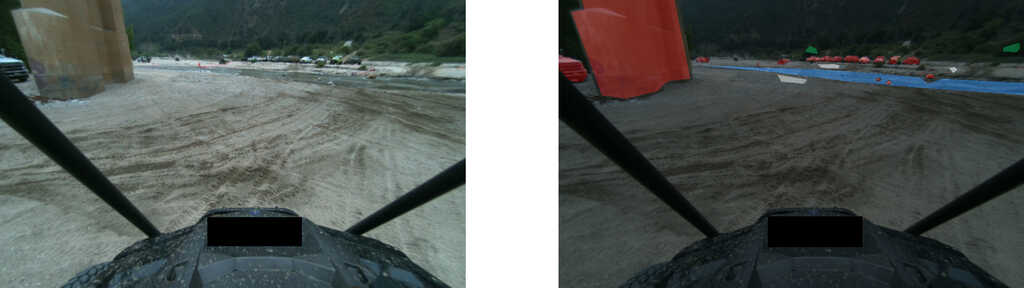}
    \includegraphics[width=\linewidth]{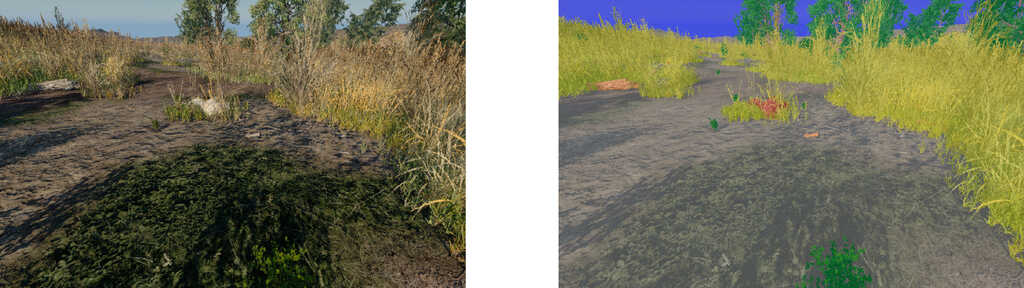}
    \caption{Samples (left) and labels (right) from our multi-biome dataset. Paso Robles Grassland (top), Mojave Desert (middle-top), San Gabriel Canyon (middle-bottom) and synthetic Forest-Sim (bottom).}
    \label{fig:custom_data}
\end{figure}

\textbf{Multi-Biome Real-World Dataset:} For the challenge of off-road autonomous driving, we have collected data in a diverse set of locations, including the Mojave Desert, Paso Robles Grassland, San Diego Grassland, California Chapparal, and San Gabriel Canyon. The data collection strategy is the same as in \cite{Atha2024FewshotSL}. This multi-biome dataset contains a large diversity of geological environments, as well as plants and water bodies. The data is collected in a ego-centric frame on a Polaris RZR vehicle. Then, 496 images are labeled with a density of less than 30\%. Additionally, we have access to around 1000 unlabeled images from each location, totaling to 5000 unlabeled samples. In addition to the real images and labels, we also generated 1000 simulations samples using the tool from Duality AI~\cite{duality}, simulating a sparse forest with tall grass, and obstacles like rocks and logs. The distribution of the simulation data differs from the real one in two aspects:  the color distribution, and the class and instances distribution. Samples of the dataset are displayed in \cref{fig:custom_data}. 

\textbf{Open-Source Off-Road Datasets} To quantify the effectiveness of our method, we consider two openly available off-road datasets: Rellis-3D \cite{Jiang2020RELLIS3DDD} and RUGD \cite{Wigness2019ARD}. The Rellis-3D data was collected at the Rellis Campus of Texas A\&M University. The environment mainly consists  of green grass, bushes, and trees. The dataset showcases high class imbalance and high ambiguity between certain classes, as is often the case for off-road datasets. The RUGD dataset is set in a semi-urban environment, across a few different locations, mainly presenting roads, parks, street lights, benches, tables, cars, rocky trails, lush vegetation, trees, and mulch. Both datasets can be mapped to a common class set, although some instances may be unique to each dataset. We choose the following class set: ground, trail, grass, water, sky, dry vegetation, lush vegetation, wall-like and diverse-obstacle. This set of classes is similar to the one used for our real-world off-road vehicle. We detail the mapping from the original class set for the two datasets in \cref{tab:class_mapping} and the class distributions in \cref{tab:class_frequencies}. 

\begin{figure}[h!]
    \centering
    \includegraphics[width=\linewidth]{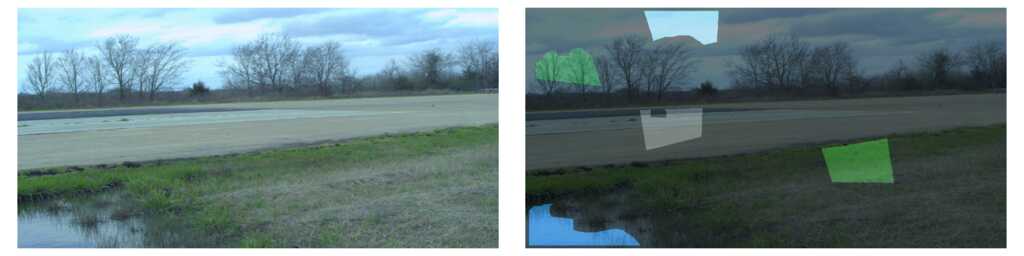}
    \includegraphics[width=\linewidth]{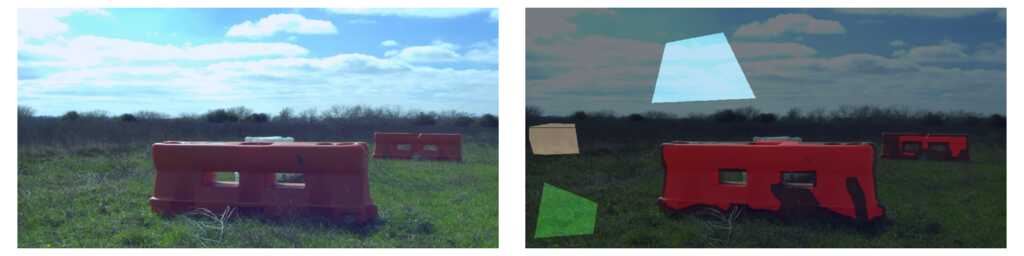}
    \includegraphics[width=\linewidth]{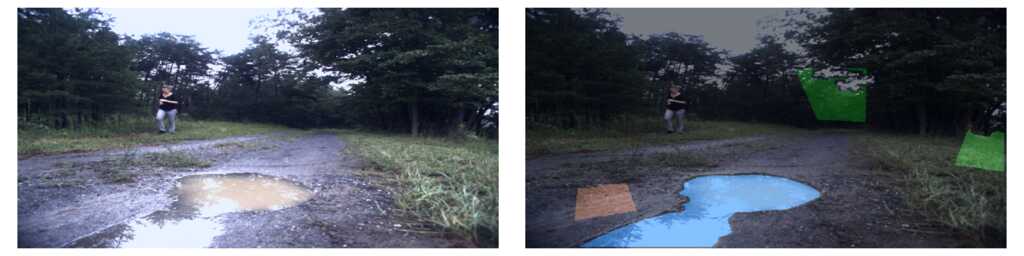}
    \includegraphics[width=\linewidth]{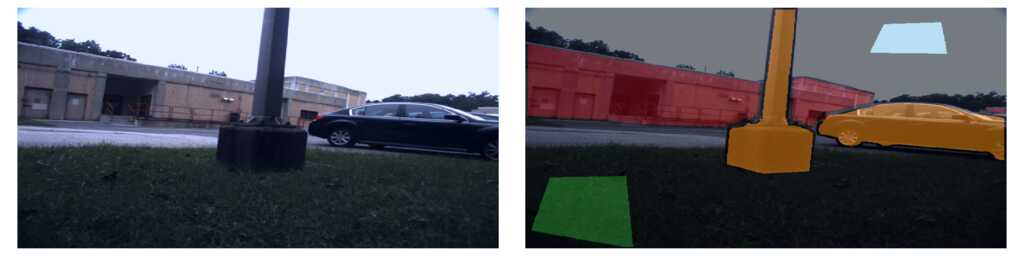}
    \caption{Samples (left) and labels (right) from the Rellis-3D dataset (top two) and RUGD (bottom two), with our custom class mapping.}
    \label{fig:rellis_rugd_data}
\end{figure}

Due to the overall cost of labeling, and especially dense labeling, even further exacerbated by the diversity of the off-road environment, we focus on the specific case of scarce, coarse, and sparse labels. Such labels are easy and inexpensive to produce, at least 1-2 images per minute. In this work, we start from the original dense labels from RUGD and Rellis-3D and sparsify according to classes. To imitate the coarse annotations in real-world off-road settings -- where annotators might quickly label broad regions rather than precise boundaries -- we remove segmentation masks in a radius of N pixels from a semantic boundary. Then we draw random polygons on the image with area equal to 1/10th the image and use the labels inside. We skip this strategy for the water, wall-like and diverse-obstacle classes which have low dataset frequency. Overall our dataset has $<20$\% label density. Samples of the coarse data are visible in \cref{fig:rellis_rugd_data}. To note, this labeling strategy is not completely representative of a real world scenario since a human annotator will most likely be aware of which classes play an important role on downstream autonomous driving performance. Nevertheless, using this data we demonstrate a lower bound on the potential performance of our method.




\subsection{Model Architecture}


We take inspiration from DPT \cite{Ranftl2021VisionTF} and design a simple custom decoder which utilizes the rich semantic features from DINOv2 and the low-level geometric features of the image to produce fine semantic boundaries. More precisely, we select features from 4 layers of the encoder, evenly spaced, similar to DPT, and fuse them in a feature neck. In parallel, we pass the image through a shallow network of convolutional blocks, with kernel sizes from 3 to 11, in a Inception-like fashion \cite{Szegedy2014GoingDW}. Finally the neck features and shallow image features are concatenated at 1/4th image resolution and passed through a fusion block, allowing to discriminate evident object boundaries. We will refer to this architecture as the PixelDecoder.

The high-level architecture is visualized in \cref{appendix:pixel_model} and detailed in \cref{appendix:pixel_decoder_architecture}. 
Note that we have not performed any ablation study to validate that the presented decoder architecture is the most appropriate for achieving the highest accuracy and generalization in semantic segmentation, but we show how it plays a role in generating accurate pseudo-labels.

\subsection{Collaborative Pseudo-Labeling}



We adopt an offline pseudo-labeling strategy to maintain greater control over the generated pseudo-labels, drawing inspiration from Cross Teaching \cite{Luo2021SemiSupervisedMI}. Our approach involves training two fundamentally distinct decoders—a CNN and a Transformer—each on different data distributions. By leveraging diverse architectures, datasets, and loss functions, these models explore distinct solution subspaces. Specifically, we employ a variant of the MaskTransformer~\cite{Strudel2021SegmenterTF}, referred to as the \textbf{PatchDecoder}, for patch-level segmentation, alongside our previously introduced \textbf{PixelDecoder}, a CNN-based decoder for pixel-level segmentation. Pseudo-labels are then derived through a disagreement procedure, discarding pixel masks where the two models’ predictions diverge. This method harnesses the strengths of both decoders, including:
\begin{itemize}
    \item The \textbf{PatchDecoder}, operating at 1/14th resolution, is robust to noisy labels and high-frequency image structures. However, its coarse resolution limits its ability to delineate precise class boundaries in lower-resolution images, making it well-suited for predicting broader, environment-level classes.
    \item The \textbf{PixelDecoder}, by contrast, excels at capturing fine-grained details and producing accurate semantic boundaries, though it can be sensitive to high-frequency image structures.
\end{itemize}

The quality of the resulting pseudo-labels depends on both the architectural differences between the decoders and the data distributions they are trained on. For best results, the distributions should be sufficiently distinct to avoid simultaneous over-confidence in the pseudo-labels, while still yielding a high density of reliable labels. To this end, we propose the following data-model pairings:

\begin{itemize}
    \item \textbf{PixelDecoder + Coarse In-Distribution + Dense Out-of-Distribution}: Given its higher-resolution predictions, the PixelDecoder benefits from learning on data with dense boundaries, which coarse in-distribution data alone cannot provide. 
    \item \textbf{PatchDecoder + Coarse In-Distribution}: The PatchDecoder, relying heavily on the encoder’s internal world representation, does not require densely labeled out-of-distribution data. Its resilience to noise also enables it to learn effectively from smaller datasets.
\end{itemize}

For example, with the target of performing semantic segmentation on RUGD, we train the PixelDecoder with the coarsely labeled samples from RUGD and the densely labeled samples from Rellis-3D, while the PatchDecoder is only trained on the coarse labels from RUGD.

This dual-decoder approach and its methodology are illustrated in \cref{fig:method}. In the rest of this work, we refer to \textbf{PatchModel} and \textbf{PixelModel} as the DINOv2 encoder with the corresponding decoder.

\section{Experiments}

\subsection{Implementation Details}

We run our experiments with the Imagenet-pretrained ViT-S/16 encoder provided by \textit{timm} \cite{rw2019timm} followed by the MaskTransformer decoder \cite{Strudel2021SegmenterTF}, SegFormer-B1 \cite{Xie2021SegFormerSA}, and DINOv2-\{S, B, L\}/14 with either a PixelDecoder, PatchDecoder.

For the coarse data, we select a subset of 300 images in each of RUGD and Rellis-3D via farthest point sampling of the CLS token embeddings produced by DINOv2-S/14. This approach ensures both repeatability and diversity of selected images. We apply the coarsification process described in~\Cref{sec:method_datasets} with a distance N=7 pixels from the semantic boundary, resulting in label densities of 7\% for Rellis and 13\% for RUGD.

We train our models for 200 epochs when using the full datasets or combining in-distribution (ID) with out-of-distribution (OOD) samples. We train for 500 epochs when using only the subselected images with coarse labels. We use the Adam optimizer with a learning rate of 0.001 and train using pixel-wise weighted cross-entropy loss. We run for 90\% of the epochs with square crops of size $512 \times 512$ and use random resizing and cropping, and finetune for 10\% of epochs at full resolution of $1024 \times 512$. To measure the overall performance of our methods, we use the mIoU metric.

\subsection{Collaborative Pseudo-Labeling}

\begin{table}[h!]
\caption{mIoU comparison across training strategies on Rellis-3D and RUGD validation datasets. ID is "in-distribution" and OOD is "out-of-distribution". We show the improvements in green versus only using the coarse ID samples. The pseudo-labels are generated using the PatchDecoder and PixelDecoder with the DINOv2-L encoder.}
\label{tab:results}
\begin{tabular}{p{0.15\linewidth} p{0.26\linewidth} p{0.2\linewidth} p{0.2\linewidth}}
\toprule
\textbf{Dataset} & \textbf{Model} & \textbf{Rellis-3D} & \textbf{RUGD} \\
\midrule
\multirow{4}{*}{\textcolor{black}{Dense ID}} \\
& \textcolor{gray}{Segmenter} & \textcolor{gray}{64.2} & \textcolor{gray}{53.3} \\
  & \textcolor{gray}{SegFormer-B1} & \textcolor{gray}{69.4} & \textcolor{gray}{60.8} \\
  & \textcolor{gray}{PatchModel (S)} & \textcolor{gray}{66.1} & \textcolor{gray}{62.8} \\
  & \textcolor{gray}{PixelModel (S)} & \textcolor{gray}{72.2} & \textcolor{gray}{67.1} \\
\midrule
\multirow{4}{*}{Coarse ID} \\
  & Segmenter & 55.1 &  43.6 \\
  & SegFormer-B1 & 56.5 & 47.5 \\
  & PatchModel (S) & 55.2 &  50.0 \\
  & PixelModel (S) & 52.1 & 47.4 \\
\midrule
\multirow{6}{1\linewidth}{Coarse ID + Dense OOD} \\
  & Segmenter & 48.8 \textcolor{BrickRed}{(-6.3)}&  45.8 \textcolor{ForestGreen}{(+2.2)} \\
  & SegFormer-B1 & 47.9 \textcolor{BrickRed}{(-8.6)}& 43.6 \textcolor{BrickRed}{(-3.9)}\\
  & PatchModel (S) & 56.5 \textcolor{ForestGreen}{(+1.3)}  &  50.3 \textcolor{ForestGreen}{(+0.3)}  \\
  & PixelModel (S) & 56.0 \textcolor{ForestGreen}{(+3.9)} & 51.1 \textcolor{ForestGreen}{(+3.7)} \\
  & \textcolor{gray}{PixelModel (B)} & \textcolor{gray}{58.2} & \textcolor{gray}{53.6} \\
  & \textcolor{gray}{PixelModel (L)} & \textcolor{gray}{59.4} & \textcolor{gray}{53.2} \\
\midrule
\multirow{4}{1\linewidth}{Pseudo-Labels} \\
  & Segmenter & 59.2 \textcolor{ForestGreen}{(+4.1)} &  48.9 \textcolor{ForestGreen}{(+5.3)}\\
  & SegFormer-B1 & 59.7 \textcolor{ForestGreen}{(+3.2)} & 54.8 \textcolor{ForestGreen}{(+7.3)}\\
  & PatchModel (S) & 58.6 \textcolor{ForestGreen}{(+3.4)} &  55.6 \textcolor{ForestGreen}{(+5.6)}\\
  & PixelModel (S) & 60.5 \textcolor{ForestGreen}{(+8.4)} & 57.1 \textcolor{ForestGreen}{(+9.7)}\\

\bottomrule
\end{tabular}
\end{table}

We establish performance baselines for our models by training them across three scenarios: dense labels, coarse labels, and a combination of in-domain coarse labels with dense out-of-domain labels. The model performance is evaluated on the validation set, with results presented in \cref{tab:results}.


We first note that the models perform worse on the RUGD dataset versus the Rellis-3D dataset, reflecting RUGD's greater complexity and diversity. We further observe that the DINOv2-S encoder provides strong features, allowing a smaller MaskTransformer decoder in the PatchModel to perform significantly better on the dense task versus a larger one in the Segmenter architecture (62.8\% vs. 53.3\%), and even a pixel-level SegFormer (60.8\%). The PixelModel benefits from leveraging both strong representations and readily available image features. 

Training solely on coarsely labeled data leads to a noticeable performance drop. On Rellis-3D the patch-level models work equivalently or better than their pixel-level counterparts, while only the PatchModel maintains the lead on the RUGD dataset. This disparity supports our previous observation of Rellis-3D displaying a much lower diversity compared to RUGD. The PixelModel, which relies heavily on image features, suffers from lack of supervised semantic boundaries.

By naively adding dense out-of-distribution data, DINOv2-based models show improvement, while others yield mixed outcomes. On the Rellis-3D, the important distributional mismatch between dense OOD data and the target coarse data limits the performance of Segmenter and SegFormer. On RUGD, the Segmenter model gets a boost from using the dense data, while SegFormer does not, effectively overfitting to Rellis' data distribution. Segmenter’s patch-level prediction capability makes it less prone to overfitting in this case. Finally, we observe that using dense labels from an out-of-distribution dataset provides a notable boost in performance for DINOv2-based models -- +1.3\% for PatchModel, +3.9\% for PixelModel on Rellis and +0.3\% for PatchModel, +3.7\% for PixelModel on RUGD. This strongly supports the idea that DINOv2 produces more domain-independent features, allowing to better transfer learned features from OOD data. 
Furthermore, scaling to larger encoders like DINOv2-B or DINOv2-L considerably improves the performance. To note, the decoder architecture was not modified to take advantage of the larger embedding dimension (see \cref{appendix:pixel_decoder_architecture} for details).

Using these trained models, we explore 4 decoder pairings -- \textbf{Patch-Patch}, \textbf{Pixel-Pixel}, \textbf{Patch-Pixel} and \textbf{Pixel-Patch} -- with three combinations of data, and find the best for pseudo-labeling:
\begin{itemize}
    \item \textbf{Coarse ID - Coarse ID}: Both decoders are trained on the same coarse in-distribution dataset. For this configuration, the first two decoder combinations are irrelevant and the last two are equal.
    \item \textbf{Dense OOD - Coarse ID}: One decoder is trained on dense out-of-distribution data and the other on coarse in-distribution data. 
    \item \textbf{Coarse ID w/ Dense OOD - Coarse ID}: Coarse in-distribution data is added to the dense out-of-distribution data
\end{itemize}

We measure the performance on the training set by quantifying the quality of the produced pseudo-labels, as displayed in \cref{tab:train_pseudo_labels}. We observe a discrepancy between the quality of pseudo-labels produced for the Rellis-3D and RUGD dataset. For Rellis-3D, the Pixel-Patch with coarse ID data combination outperforms the case where a model is trained only on dense OOD data. Furthermore, the Pixel-Pixel (S) model combination leads the other models in their respective data configurations. On the other hand, the Pixel-Patch (S) model combination generates the highest quality pseudo-labels for the RUGD dataset. The difference in optimal model pairings is likely related to Rellis-3D’s lower diversity and noisier labels \cite{Atha2024FewshotSL} compared to RUGD (see \cref{tab:class_frequencies} for data distribution). As such, we find that the results on the RUGD data may be more representative of our desired use case, i.e., using dense out-of-distribution (simulation) labels to enable better performance across a range of environments. Tangentially, we note that using a larger encoder further boosts the quality of pseudo-labels, leading us to use the DINOv2-L/14 encoder for downstream pseudo-label generation.

Finally, using our proposed method, we train and evaluate the segmentation models on the generated pseudo-labels (see results in \cref{tab:results}). We find a significant boost in performance over using only the coarse labels, or using the naive mixing of coarse and dense data. This improvement is seen across all models, and not just the DINOv2-based ones. Our PixelModel achieves an 8.4\% and 9.7\% mIoU improvement compared to coarse-label-only training on Rellis-3D and RUGD, respectively.

\begin{table}[h]
\centering
\caption{mIoU comparison on the Rellis-3D and RUGD training datasets. ID is "in-distribution" and OOD is "out-of-distribution". The configurations indicate the dataset used for each model, e.g., for the first run, both models use the coarse ID dataset. We underline the best model combination with the small encoder and highlight the best overall model in bold.}
\label{tab:train_pseudo_labels}
\begin{tabular}{p{0.25\linewidth} p{0.28\linewidth} p{0.15\linewidth} p{0.15\linewidth}}
\toprule
\textbf{Configuration} & \textbf{Models} & \textbf{Rellis-3D} & \textbf{RUGD} \\
\midrule
\multirow{1}{1\linewidth}{\textit{Coarse ID - Coarse ID}} \\

& Pixel-Patch (S) & 72.1 & 70.3 \\
\midrule
\multirow{3}{1\linewidth}{\textit{Dense OOD - Coarse ID}} \\
& Patch-Patch (S) & 65.6 & 65.0 \\
& Pixel-Pixel (S)& 71.0 & 68.3 \\
& Patch-Pixel (S) & 68.2 & 65.0 \\ 
& Pixel-Patch (S) & 68.9 & 71.5 \\ 
\midrule
\multirow{5}{1\linewidth}{\textit{Coarse ID w/ Dense OOD - Coarse ID}} \\
& Patch-Patch (S) & 73.6 & 71.6 \\
& Pixel-Pixel (S)& \underline{75.7} & 71.0 \\
& Patch-Pixel (S) & 75.6 & 72.4 \\ 
& Pixel-Patch (S) & $74.1$ & \underline{73.3} \\ 
& Pixel-Patch (B) & $76.1$ & $73.5$ \\ 
& Pixel-Patch (L) & \textbf{77.3} & \textbf{73.8} \\
\bottomrule
\end{tabular}
\end{table}

\subsection{Collaborative Semantic Segmentation for Off-Road Au-
tonomous Driving}

We deployed COARSE on our custom autonomous vehicle platform, a Polaris RZR outfitted with an extensive sensor suite -- including LiDARs, radars, RGB and stereo cameras—designed to enable off-road navigation in GPS-denied environments across a wide range of biomes.

As detailed in \cref{sec:method_datasets}, our trials spanned diverse biomes: the Mojave Desert, Paso Robles Grassland, San Diego Grassland, California Chaparral and San Gabriel Canyon. The Mojave Desert, with its dry bush and ground, posed a simpler challenge, though distinguishing trails from ground proved tricky -- correct trail prediction being critical for enabling higher speeds. Paso Robles Grassland featured yellow grass, tree canopies, branches, and logs, with significant illumination shifts between forested and open areas. San Diego Grassland’s tall, sight- and LiDAR-impermeable grass tested navigation limits. The San Gabriel, with frequent river crossings, challenged water detection across reflective, murky, foamy, and transparent forms. To generate pseudo-labels, we leveraged dense out-of-distribution simulation data from the Forest Sim. To tackle the lack of some real-world classes in the simulation data, we followed a simple cut-and-paste procedure in order to introduce some classes to the models during training.

\begin{figure}[h!]
    \centering
    \includegraphics[width=1\linewidth]{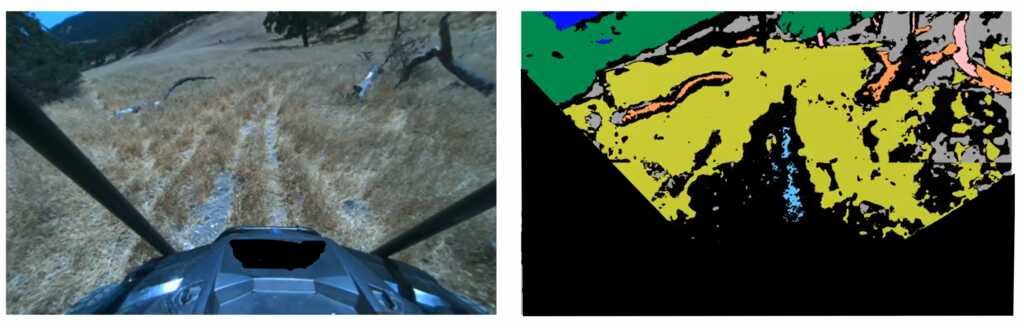}
    \includegraphics[width=1\linewidth]{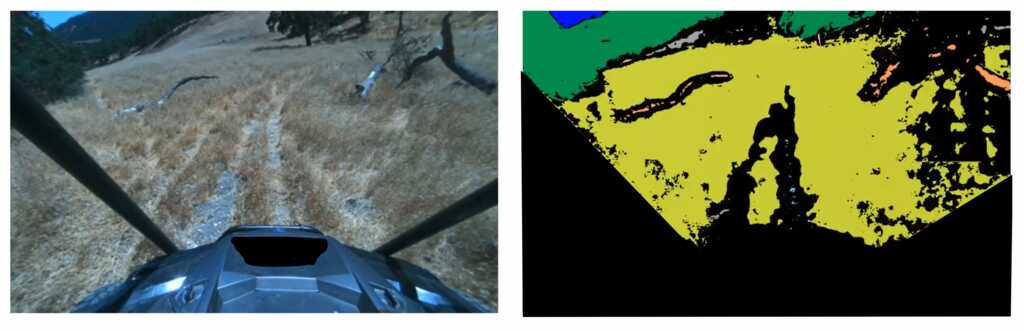}
    \includegraphics[width=1\linewidth]{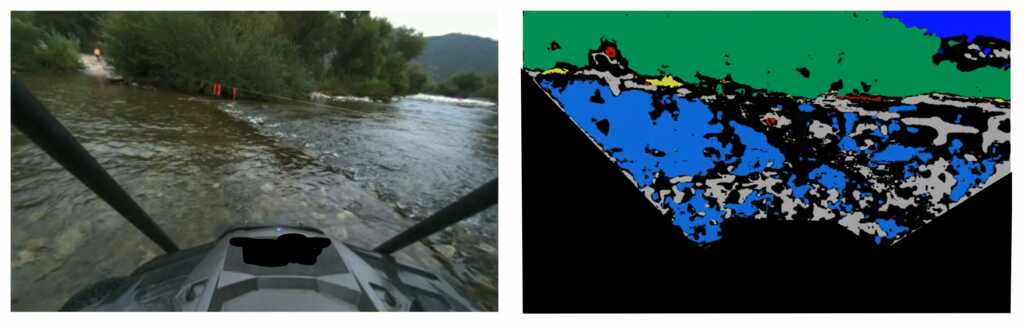}
    \includegraphics[width=1\linewidth]{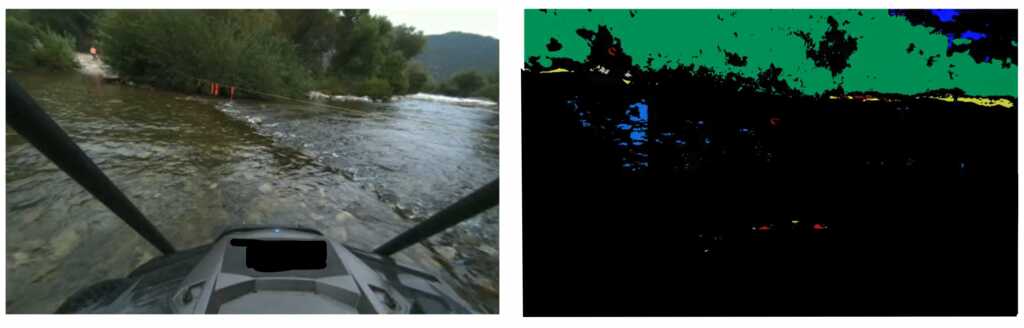}
    \caption{Images (left) and pseudo-labels (right) for the Paso Robles Grassland (top two) and Sang Gabriel Canyon (bottom two). The first and third samples are generated with a Pixel-Pixel model combination, while the second and fourth are generated with the Pixel-Patch pairing.}
    \label{fig:racer_compare_pl}
\end{figure}

Using our custom data split of dense simulation labels + coarse real labels and the coarse real labels alone, we show generated pseudo-labels for the Pixel-Pixel and Pixel-Patch model combinations on the Paso Robles Grassland and the San Gabriel Canyon in \cref{fig:racer_compare_pl}. For the San Gabriel Canyon, accurate labeling of water (dark blue) was primordial for stable water crossings. We observe that the pseudo-labels which use the same Pixel-level decoder are very confident, yielding inaccurate prediction of the water as ground (gray). Alternatively, the pseudo-labels produced by the COARSE method show a large sparsity around water, avoiding interjecting noise in the labels.  Similarly we observe overconfidence in the Paso Robles Grassland, where the top part of a log (orange) is predicted as a trunk (pink), and grass (yellow) is predicted as ground. The sparsity of generated pseudo-labels is proportional to both the difficulty and level of semantic confusion. Labeling of diverse and complex images could therefore be guided by the simple heuristic of pseudo-label density.


\section{Conclusion}

We introduced COARSE, a novel semi-supervised domain adaptation framework tailored for off-road semantic segmentation, addressing the critical challenge of scarce, coarsely labeled data in unstructured environments. By leveraging the robust feature extraction capabilities of pretrained vision transformers like DINOv2 and integrating complementary PixelDecoder and PatchDecoder architectures, COARSE effectively bridges domain gaps and enhances segmentation performance. Our collaborative pseudo-labeling strategy, which combines coarse in-domain labels with dense out-of-domain data, produces high-quality pseudo-labels that significantly boost model generalization.

COARSE offers a scalable and cost-effective solution to the labeling bottleneck in off-road autonomy, reducing reliance on expensive, dense annotations while harnessing the abundance of unlabeled data and simulated environments. The pseudo-labeling approach not only improves performance but also provides a heuristic based on density for identifying areas needing further annotation, paving the way for iterative refinement in real-world applications. Looking forward, we aim to explore other ways to exploit the generalization capabilities of the DINOv2 foundational model to develop offline and online pseudo-labeling strategies.



\bibliographystyle{IEEEtran}

\bibliography{refs}

\appendix



\begin{table*}[h!]
    \caption{Mapping of Custom Dataset Classes to Rellis-3D and RUGD Classes}
    \label{tab:class_mapping}
    \centering
    \begin{tabular}{l p{0.3\textwidth} p{0.3\textwidth} p{0.1\textwidth}}
        \toprule
        \textbf{Custom Dataset Class} & \textbf{Rellis-3D Classes} & \textbf{RUGD Classes} \\
        \midrule
        ground & dirt, mud & dirt, sand, mulch, rock-bed & \cellcolor[RGB]{139,69,19} \\
        trail & asphalt, concrete & asphalt, gravel, concrete & \cellcolor[RGB]{105,105,105} \\
        grass & grass & grass & \cellcolor[RGB]{34,139,34} \\
        water & water, puddle & water & \cellcolor[RGB]{30,144,255} \\
        sky & sky & sky & \cellcolor[RGB]{135,206,235} \\
        dry\_vege & bush & bush & \cellcolor[RGB]{222,184,135} \\
        lush\_vege & tree & tree & \cellcolor[RGB]{0,100,0} \\
        wall-like & building, fence, barrier & building, fence, bridge & \cellcolor[RGB]{139,0,0} \\
        diverse-obstacle & pole, vehicle, object, person, log, rubble & pole, vehicle, object, sign, rock, picnic-table, bicycle, person, log & \cellcolor[RGB]{255,140,0} \\
        \bottomrule
    \end{tabular}
    
\end{table*}

\begin{table}[h]
\centering
\caption{Class Frequencies for Rellis-3D and RUGD Datasets}
\label{tab:class_frequencies}
\begin{tabular}{lll}
\toprule
\textbf{Class} & {\textbf{Rellis-3D (\%)}} & {\textbf{RUGD (\%)}} \\
\midrule
ground          &  2.86 & 11.71 \\
trail           &  1.06 & 10.95 \\
grass           & 33.59 & 23.85 \\
water           &  0.65 &  0.11 \\
sky             & 30.02 &  8.22 \\
dry\_vege       & 15.83 &  2.37 \\
lush\_vege      & 15.03 & 39.33 \\
wall-like       &  0.50 &  2.02 \\
diverse-obstacle &  0.46 &  1.45 \\
\bottomrule
\end{tabular}
\end{table}


\begin{figure*}[h!]
    \centering
    \includegraphics[width=\linewidth]{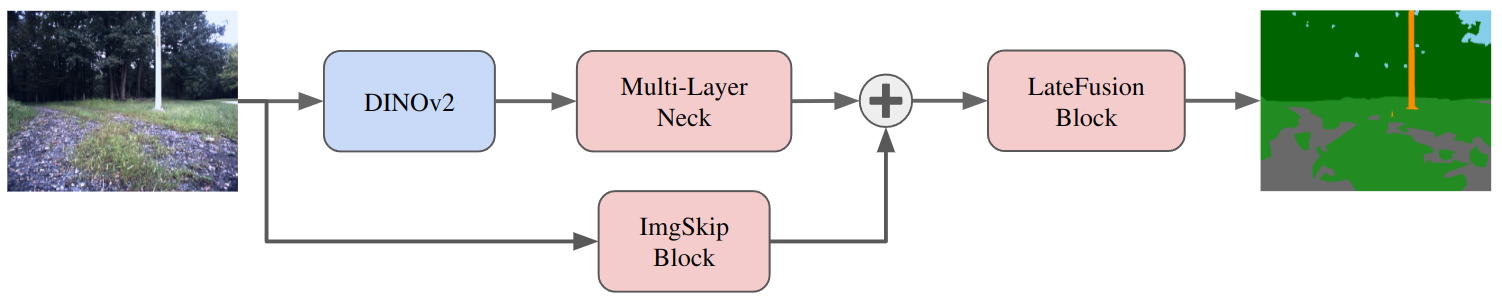}
    \caption{Architecture of the PixelModel. The DINOv2 encoder produces rich semantic features from the image which are fused in the Multi-Layer Neck. The image also bypasses the encoder through a shallow CNN, the ImgSkip Block. The high-level semantic features are fused with the low-level geometric features at 1/4th the image resolution in the LateFusion Block to produce the final prediction.}
    \label{appendix:pixel_model}
\end{figure*}

\begin{table}[h]
\centering
\caption{Comparison of Model Parameters, Trainable Parameters, and Inference Speed on an H100 GPU}
\label{tab:model_parameters_fps}
\begin{tabular}{lp{0.22\linewidth}p{0.22\linewidth}p{0.1\linewidth}}
\toprule
\textbf{Model} & {\textbf{Total Parameters (M)}} & {\textbf{Trainable Parameters (M)}} & {\textbf{FPS}} \\
\midrule
Segmenter     & 25.8 & 4.0 & 105.65 \\
SegFormer-B1  & 13.7 & 0.5 & 184.34 \\
PatchModel    & 24.3 & 2.2 &  77.74 \\
PixelModel    & 22.7 & 0.7 &  68.10 \\
\bottomrule
\end{tabular}
\end{table}

\begin{table}[h]
\centering
\caption{Summary of PixelDecoder Architecture. $d$ is the encoder embedding dimension. FF\# refers to encoder layer where the features are taken from.}
\label{appendix:pixel_decoder_architecture}
\begin{tabular}{p{0.2\linewidth}p{0.2\linewidth}p{0.2\linewidth}p{0.2\linewidth}}
\toprule
\textbf{Layer} & \textbf{Input Channels} & \textbf{Output Channels} & \textbf{Kernel Size} \\
\midrule
FF3 & d & 64  & (1, 1) \\
FF6 & d & 128 & (1, 1) \\
FF9 & d & 192 & (1, 1) \\
FF12 & d & 256 & (1, 1) \\
\midrule
Fuser & 640 & 128 & (1, 1) \\
\midrule
Compressor & 128 & 64  & (1, 1) \\
\midrule
\multirow{5}{0.1\linewidth}{ImgSkip Preprocessor} & 3 & 9 & (3, 3) \\
                                              & 3 & 9 & (5, 5) \\
                                              & 3 & 9 & (7, 7) \\
                                              & 3 & 9 & (9, 9) \\
                                              & 3 & 9 & (11, 11) \\
\midrule
\multirow{2}{0.1\linewidth}{ImgSkip Fuser} & 45 & 64 & (3, 3) \\
                                          & 64 & 64 & (3, 3) \\
\midrule
\multirow{4}{0.1\linewidth}{LateFusion Block} & 128 & 64 & (5, 5) \\
                      & 64  & 64 & (3, 3) \\
                      & 64  & 64 & (3, 3) \\
                      & 64  & 9  & (1, 1) \\
\bottomrule
\end{tabular}
\end{table}




\end{document}